\documentclass[letterpaper]{article} 
\usepackage{aaai24}  
\usepackage{times}  
\usepackage{helvet}  
\usepackage{courier}  
\usepackage[hyphens]{url}  
\usepackage{graphicx} 
\usepackage{natbib}  
\usepackage{caption} 
\usepackage{algorithm}
\usepackage{algorithmic}
\usepackage{amsmath}
\usepackage{booktabs}
\usepackage{amssymb}
\usepackage{epsfig}
\usepackage{graphicx}
\usepackage{booktabs}
\usepackage{bbding}
\usepackage{subfigure}
\usepackage{cuted}

\usepackage{newfloat}
\usepackage{listings}
\DeclareCaptionStyle{ruled}{labelfont=normalfont,labelsep=colon,strut=off} 
\floatstyle{ruled}
\newfloat{listing}{tb}{lst}{}
\floatname{listing}{Listing}
\title{TagCLIP: A Local-to-Global Framework to Enhance Open-Vocabulary Multi-Label Classification of CLIP Without Training}
\author{
    Yuqi Lin\textsuperscript{\rm 1,3},
    Minghao Chen\textsuperscript{\rm 2}\thanks{Corresponding authors},
    Kaipeng Zhang\textsuperscript{\rm 3}\footnotemark[1],
    Hengjia Li\textsuperscript{\rm 1},
    Mingming Li\textsuperscript{\rm 1},\\
    Zheng Yang\textsuperscript{\rm 4},
    Dongqin Lv\textsuperscript{\rm 6},
    Binbin Lin\textsuperscript{\rm 5},
    Haifeng Liu\textsuperscript{\rm 1},
    Deng Cai\textsuperscript{\rm 1,4}
}
\affiliations{
    \textsuperscript{\rm 1}State Key Lab of CAD\&CG, College of Computer Science, Zhejiang University\\
    \textsuperscript{\rm 2}Hangzhou Dianzi University
    \textsuperscript{\rm 3}Shanghai Artificial Intelligence Laboratory
    \textsuperscript{\rm 4}FABU Inc. \\
    \textsuperscript{\rm 5}School of Software Technology, Zhejiang University \;
    \textsuperscript{\rm 6}Nantong Port Group

    \{linyq5566, minghaochen01\}@gmail.com, \; kp\_zhang@foxmail.com
}

\usepackage{bibentry}

\begin{document}

\maketitle

\begin{abstract}
Contrastive Language-Image Pre-training (CLIP) has demonstrated impressive capabilities in open-vocabulary classification. The class token in the image encoder is trained to capture the global features to distinguish different text descriptions supervised by contrastive loss, making it highly effective for single-label classification. However, it shows poor performance on multi-label datasets because the global feature tends to be dominated by the most prominent class and the contrastive nature of softmax operation aggravates it.
In this study, we observe that the multi-label classification results heavily rely on discriminative local features but are overlooked by CLIP. As a result, we dissect the preservation of patch-wise spatial information in CLIP and proposed a local-to-global framework to obtain image tags. It comprises three steps: (1) patch-level classification to obtain coarse scores; (2) dual-masking attention refinement (DMAR) module to refine the coarse scores; (3) class-wise reidentification (CWR) module to remedy predictions from a global perspective. 
This framework is solely based on frozen CLIP and significantly enhances its multi-label classification performance on various benchmarks without dataset-specific training. Besides, to comprehensively assess the quality and practicality of generated tags, we extend their application to the downstream task, i.e., weakly supervised semantic segmentation (WSSS) with generated tags as image-level pseudo labels. Experiments demonstrate that this \textit{classify-then-segment} paradigm dramatically outperforms other annotation-free segmentation methods and validates the effectiveness of generated tags. Our code is available at \url{https://github.com/linyq2117/TagCLIP}.
\end{abstract}

\begin{figure}[t]
  \centering
   \includegraphics[width=0.96\linewidth]{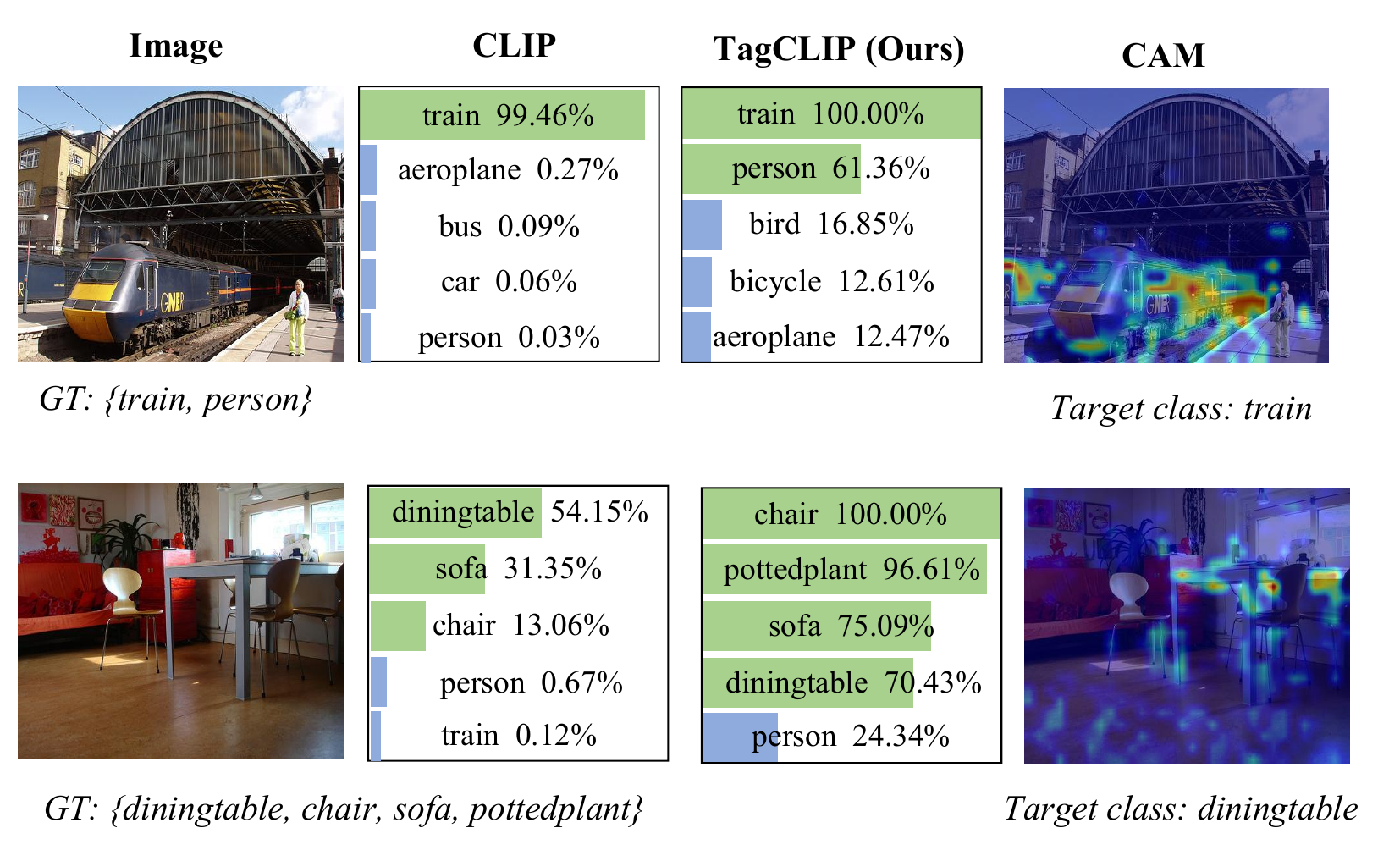}
   \caption{Visualizations of multi-label classification results and CAMs of some target classes. The middle two columns demonstrate that original CLIP~\cite{CLIP} usually fails to recognize inconspicuous categories while our TagCLIP can identify them well. The last column presents some CAMs of specific classes and indicates that classification mainly depends on some discriminative local features. All results are based on ViT-B/16 and we leverage GradCAM~\cite{gradcam} to obtain CAMs for CLIP.}
   \label{fig:1}
\end{figure}

\section{Introduction}
Contrastive Language-Image Pre-training (CLIP)~\cite{CLIP} has recently emerged as a powerful vision-language model. 
It is pre-trained on a large-scale dataset of image-text pairs and has shown impressive performance in image-text matching tasks~\cite{zhou2022coop, crowson2022vqgan, gu2021opendet}. By transferring this matching ability to the classification task, we can recognize arbitrary text labels and achieve open-vocabulary classification. However, most existing open-vocabulary works focus on the single-label classification task while multi-label classification, which aims to recognize all the relevant categories or concepts in an image, is a more practical and challenging task. In Figure~\ref{fig:1}, we find that the performance is unsatisfactory on multi-label classification datasets. Specifically, the classification logits predicted by the class token tend to be dominated by the most prominent class, while some inconspicuous objects, e.g., with small size, are usually underrated. It stems from two main reasons: (1) CLIP is trained to align image-text pairs with contrastive loss, which aims to match an image with its corresponding text descriptions and distinguish it from others. The softmax operation introduced by this loss creates competition among different classes, which is detrimental to the multi-label setting. 
(2) CLIP is trained to represent an entire image through a unique global embedding using the class token, without explicitly capturing the local features of specific regions. However, in the multi-label setting, discriminative local features are more helpful. This preference for local features can be observed in the Class Activation Map (CAM)~\cite{cam} shown in Figure\ref{fig:1}, where the highly responsive regions for the target class mainly correspond to specific local cues.
Therefore, it is necessary to explore the spatial information preserved in CLIP-ViT to take advantage of discriminative local cues.

In general, the final output feature map of a model is commonly utilized for localization tasks, e.g., object detection~\cite{ren2015fasterrcnn} or segmentation~\cite{chen2017deeplab}. However, we observe that the localization quality of CLIP-ViT is not effective for the last feature map~(seeing Figure~\ref{fig:spatial}). We delve into the underlying factors and find that the attention operation in the last layers is irrational for dense tokens, leading to the lack of spatial information in the final output feature map. Alternatively, by forwarding the feature map of the penultimate layer without the self-attention operation at the last layer~(denoted as penultimate layer for short), the spatial information is effectively preserved. This enables us to extract local features from CLIP, enhancing its capability for capturing fine-grained details.

Building on the observation above, we further propose a novel framework called TagCLIP to enhance the multi-label classification capability of the original CLIP without training. This framework follows a local-to-global paradigm and consists of three steps. First, we ignore the attention operation at the last layer of CLIP-ViT and perform patch-level classification based on the penultimate layer to obtain corresponding classification score maps for each class. Second, to refine the initial scores and mitigate potential noise, we introduce a dual-masking attention refinement strategy based on the Multi-Head Self-Attention (MHSA) inherent in ViT. 
Finally, we propose a class-wise reidentification module to further improve the primary predictions from the global view. This double-check approach can filter out some falsely detected classes and improve the scores of missed cases. 
The whole framework remarkably improves the multi-label classification performance of CLIP. It is based solely on frozen CLIP and enables open-vocabulary multi-label classification without the need for dataset-specific training.

To further validate the quality and practicality of generated tags, we integrate TagCLIP with downstream tasks, where it serves as a generalizable annotator that provides high-quality pseudo labels. It can benefit many downstream tasks, e.g., self-training~\cite{zoph2020rethinkingselftraining, wang2022debiased, xie2020selftraining}, and weakly supervised learning~\cite{lin2022clipes, Xie_2022_CLIMS, xu2022mctformer}. In this paper, we explore the application of TagCLIP by integrating the generated labels with weakly supervised semantic segmentation (WSSS). The combination of open-vocabulary multi-label classification and WSSS enables annotation-free segmentation. Unlike previous works~\cite{zhou2022maskclip, van2021maskcontrast} following the bottom-up paradigm, we surprisingly find this novel \textit{classify-then-segment} paradigm leads to significant performance gains, which indicates the importance of image-level supervision to the segmentation task.

The main contributions can be summarized as follows:
\begin{itemize}
\item We explore the spatial information in CLIP at the patch level and find that the attention operation in the last layer breaks spatial information. On this basis, we propose a local-to-global framework TagCLIP to enhance the multi-label classification performance of the original CLIP without any extra training.

\item Experiment results demonstrate the effectiveness of our TagCLIP. It unlocks the potential of original CLIP and can generate high-quality image tags. 
Our method achieves significant performance gains compared to original CLIP and other works across different benchmarks.

\item We integrate the proposed TagCLIP with the downstream WSSS task and find this \textit{classify-then-segment} paradigm achieves remarkable improvement over other methods.
\end{itemize}

\section{Related Works}

\subsection{Contrastive Language-Image Pre-training}
Contrastive Language-Image Pre-training(CLIP)~\cite{CLIP} connects visual concepts with textual descriptions and has empowered many computer vision tasks with language ability. It consists of an image and text encoder, and is jointly trained to align the two modalities with over 400 million image-text pairs. The image-text matching ability can be transferred to the downstream zero-shot tasks. However, the pre-training task is image-level, and only \textit{class} token is trained to capture the global feature. For multi-label classification task, the region-level feature is preferred. 
Some works~\cite{raghu2021dovision, ghiasi2022whatdovision} explore the spatial information in the patches of deep ViT layers but the results are unsatisfactory. This paper makes it possible by ignoring the last attention operation, and leverages obtained local features to benefit multi-label classification.

\begin{figure*}
  \centering
  \includegraphics[width=0.94\linewidth]{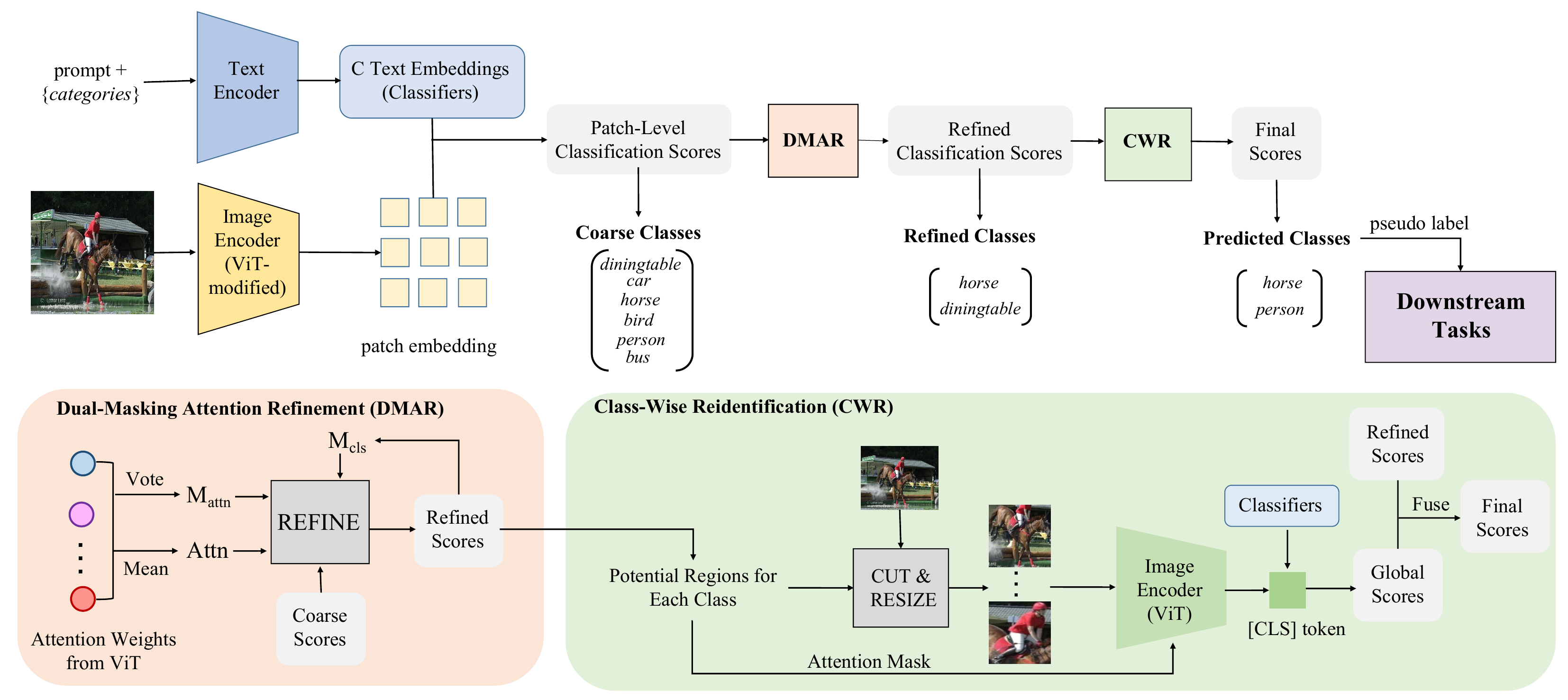}
  \caption{An overview of our proposed framework. The framework consists of three steps, i.e., patch-level classification, dual-masking attention refinement (DMAR), and class-wise reidentification (CWR). \textit{C} is the total number of classes. ``ViT-modified" means ignoring the last self-attention operation to maintain spatial information. We threshold the predicted probability scores with 0.5 to obtain predicted classes. The predicted image tags can be treated as pseudo labels for downstream tasks, e.g., WSSS.}
  \label{fig:framework overview}
\end{figure*}

\subsection{Open-Vocabulary Multi-Label Classification}
Multi-Label Classification aims to predict a set of labels for an image. Conventionally, a multi-label classification task is transformed into a set of binary classification tasks, which are solved by optimizing a binary cross-entropy loss function. The proposed methods can be categorized into three main directions: 1) Improving loss functions~\cite{ridnik2021asymmetric, wu2020distribution}. 2) Modeling label correlations~\cite{chen2019multi, chen2019learning, ye2020attention}. 3) Locating regions of interest~\cite{wang2017multi, you2020cross}. To deal with unseen labels, multi-label zero-shot learning (ML-ZSL) is developed to transfer knowledge from seen classes to unseen classes. The keys to this task are the alignment of the image with its relevant label embeddings and the relation between seen and unseen label embeddings. Existing works realize it from the perspective of finding principle directions~\cite{ben2021zs-sdl} or adopting attention module~\cite{narayan2021bira,huynh2020lesa}. 

Different from ML-ZSL, visual-related language data like image captions can be used as auxiliary supervision in the open vocabulary setting. The open-vocabulary multi-label recognition can classify multi-label images via arbitrary textual names or descriptions. According to the complexity, existing methods can be divided into two groups. 1) The first group requires additional training processes on seen classes or specific curated data. These methods require fine-tuning on target datasets~\cite{he2023MKT, sun2022dualcoop} or training from scratch using massive data~\cite{guo2023TAI}, both of which have complex training processes. 2) The second group is merely based on pre-trained models without further training or extra information~\cite{li2023clipsurgery}. Our work falls into the second group, which is more challenging but convenient to use. Similar to CLIP-Surgery~\cite{li2023clipsurgery}, we improve the classification ability of CLIP from the perspective of model explainability. The difference is that we leverage a local-to-global framework, while CLIP-Surgery only relies on global embedding.

\subsection{Annoation-Free Semantic Segmentation}
In the annotation-free segmentation setting, no annotation is provided during training, which is corresponding to unsupervised semantic segmentation~(USS).
Primary USS methods leverage self-supervised learning to learn pixel-level representation~\cite{ji2019iic,cho2021picie,ziegler2022leopart,ke2022hsg,hwang2019segsort,van2021maskcontrast} and the learned representations can then be employed to cluster image segments via K-means or linear classifiers. These bottom-up approaches are difficult to distinguish different classes with similar appearances or identify classes with varied appearances. 

Another similar setting is open-vocabulary segmentation. Its target is to segment an image with arbitrary categories described by texts instead of fixed labeling vocabularies. It typically addresses the closed-set limitation via training on weak supervision signals, e.g., image-text pairs~\cite{xu2022groupvit, luo2023segclip}. Differently, recent works~\cite{zhou2022maskclip,shinreco,shin2022namedmask} are merely based on pre-trained CLIP and require no extra annotations. The performance gain of these methods is still limited for the lack of high-level semantic guidance. We denote all the above methods leveraging image-text pairs or the pre-trained model as CLIP-based methods.

\section{Method}
In this section, we introduce our CLIP-based multi-label classification framework, TagCLIP, which is depicted in Figure~\ref{fig:framework overview}. We first review the architecture of CLIP-ViT and investigate the spatial information preserved in the patches. Then, we introduce the proposed local-to-global framework for multi-label classification without annotations and finetuning. Finally, we present the application of generated image tags on the downstream WSSS task.

\subsection{Analysis of CLIP}
CLIP~\cite{CLIP} consists of an image encoder and a text encoder and is jointly trained to align the two modalities with large-scale image-text pairs. For the image encoder with transformer architecture, a [\textit{cls}] token is pre-trained to capture the global feature. Given the ViT with $L$ layers, the forward propagation of the last transformer layer is expressed as follows:
\begin{align}
  \hat{X}^{L} &= X^{L-1} + \mathbf{a}^{L}, \\
  &= X^{L-1} + A^{L}\left(X^{L-1} W_V^{L}\right), \\
  A^{L} &= \sigma (\frac{(X^{L-1} W_Q^{L})(X^{L-1} W_V^{L})^T}{\sqrt{d}} + M^{L}),
\label{eq123}
\end{align}

\begin{equation}
  X^L = \hat{X}^{L} + \operatorname{MLP}(\hat{X}^{L}),
\label{eq4}
\end{equation}
where $X^{L-1}$ represents the output tokens of the \textit{L-1} layer, $\mathbf{a}^{L}$ and $\operatorname{MLP}$ represent the self-attention and the MLP modules in the transformer block. $A^{L}$ encodes the attention weights at layer $L$. $\sigma$ represents the softmax normalization, $d$ is the dimension of $X^{L-1}$, $M^{L}$ is attention mask for $A^{L}$. $W_Q, W_K, W_V$ are linear projection weights to generate \textit{query, key, value} in MHSA. $X^L$ consists of the [\textit{cls}] token and remaining tokens (denoted as dense tokens):
\begin{equation}
  X^L = [x^L_{cls}, x^L].
  \label{eq5}
\end{equation}

As mentioned in Introduction, the contrastive loss and global embedding in the original CLIP will harm the multi-label classification. Alternatively, region-level features are better suited to recognize multiple categories in an image. As only the [\textit{cls}] token is used during contrastive pre-training, the localization ability of the original CLIP is weak~\cite{zhong2022regionclip}. There is a major performance degradation when applying the pre-trained CLIP model for localization tasks~(e.g., only 16.2\% mIoU for segmentation by leveraging the final output feature map in Table~\ref{tab:without last attn}). 

We hypothesize that the spatial information is remained in feature maps of CLIP in the previous layer but lacks in the last layer for the following reasons: (1) The query and key in the last attention layer are merely involved in the optimization of [\textit{cls}] token to perform weighted sum operation and globalizes information during pre-training. It is a special design for [\textit{cls}] token but is meaningless and redundant for remaining dense tokens. (2) The [\textit{cls}] token plays a relatively minor role throughout the vision transformer and is not used for globalization until the last layer~\cite{ghiasi2022whatdovision}. Therefore, it scarcely affects local features in previous layers. To verify it, we use ViT-B/16 with 12 layers and treat the encoded text features as classifiers to classify each dense token outputted by the last two layers. To make the feature embedded into the same feature space, we let the dense token outputted by the penultimate layer pass the rest layer without self-attention:
\begin{align}
  \hat{x}_{dense} &= x^{L-1} + \mathbf{c}^{L}, \\
  &= x^{L-1} + x^{L-1} W_V^{L}, \\
  x_{dense} &= \hat{x}_{dense} + \operatorname{MLP}(\hat{x}_{dense}).
\label{eq678}
\end{align}

\begin{figure}[t]
  \centering
   \includegraphics[width=0.8\linewidth]{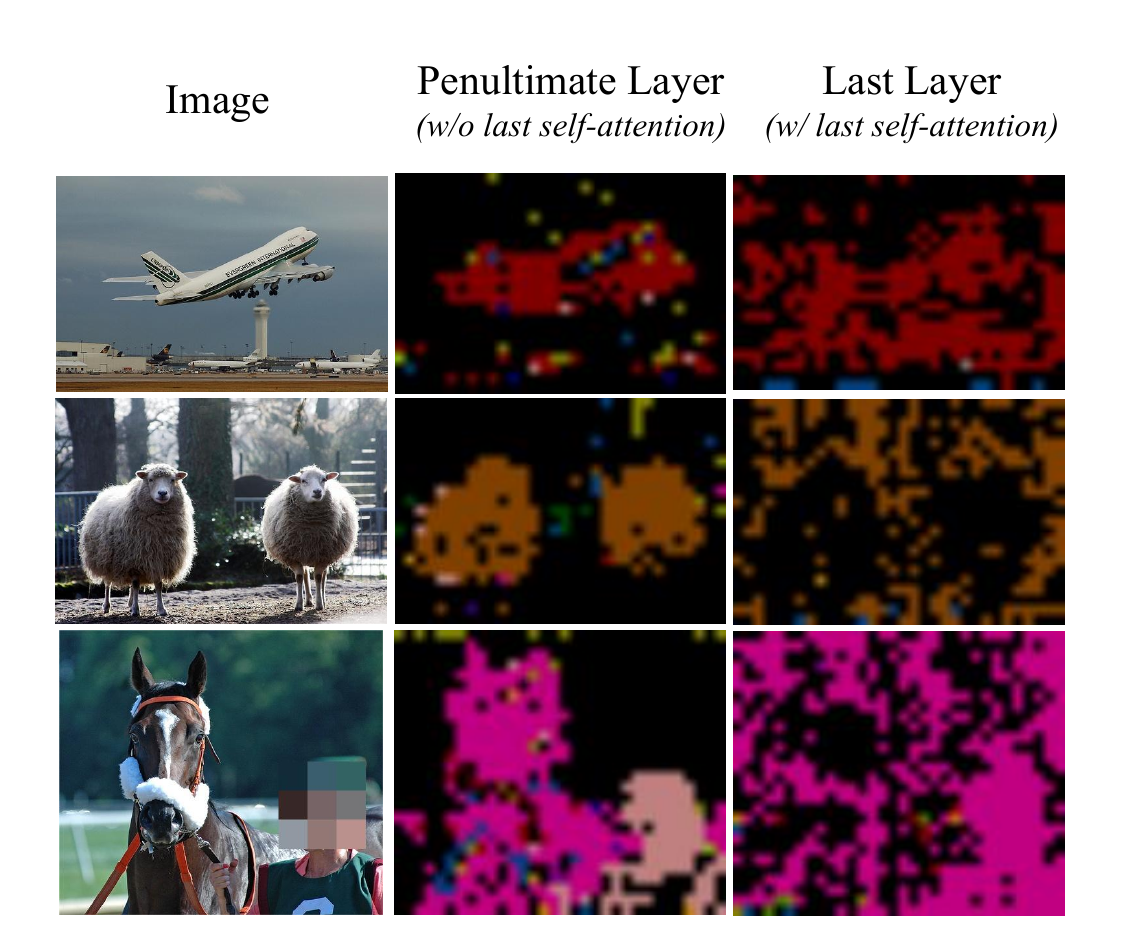}
    \vspace{-2mm}
   \caption{Qualitative results of the patch-level classification upon $x_{dense}$ and $x^L$ outputted by the last two layers of CLIP-ViT respectively. The last self-attention operation breaks spatial information in ViT. We blurred the human face for the ethical consideration.
   }
   \label{fig:spatial}
\end{figure}

We provide qualitative and quantitative results in Figure~\ref{fig:spatial} and Table~\ref{tab:without last attn}. The results demonstrate spatial information is preserved in the penultimate layer but lacks in the last layer. Therefore, it is feasible to omit the last self-attention operation and perform classification based on the projected output of the penultimate layer to discover the discriminative features for target classes.

\subsection{CLIP-Based Multi-Label Classification}
This section introduces our proposed local-to-global framework for multi-label classification, including patch-level classification to obtain coarse scores, dual-masking attention refinement (DMAR) to refine coarse scores, and the class-wise reidentification (CWR) module to double-check the potential predictions. 

\begin{table}
  \begin{center}
  \begin{tabular}{ccc}
    \toprule
    Last Self-Attention  & mAP   &  mIoU  \\
    \midrule
    \Checkmark      &    82.7  & 16.2 \\
    \XSolidBrush     &      85.4  & 41.6 \\
    
    \bottomrule
  \end{tabular}
  \end{center}
  \caption{Quantitative results for the effect of last self-attention operation in terms of classification~(mAP) and segmentation~(mIoU) on PASCAL VOC 2012 validation set.}
  \label{tab:without last attn}
  
\end{table}

\subsubsection{Coarse Classification}
To perform patch-level classification, the output feature map based on the penultimate layer $x_{dense} \in R^{N \times D}$ is leveraged. The output of the text encoder is denoted as $T \in R^{D \times C}$, which acts as the classifier based on the text inputs. $N, D, C$ represent token length, token dimension and class number, respectively. The classification score for each patch in $x_{dense}$ is calculated as:
\begin{equation}
  s_{i} = Linear(x_{dense,i}) * T,
  \label{eq9}
\end{equation}
where $i$ represents the spatial index of each patch. $Linear$ is the last layer of CLIP to map the encoded image features and text features into the unified space in CLIP. $s_{i}$ reflects the similarity of image token and $C$ text descriptions, and the similarity scores will be forwarded to the softmax function to normalize these scores over all classes (Note that the softmax operation is optional, but we find it significant for CLIP and validate it in the experiment section). The probability classification score of class $c$ for each dense token $i$ can be obtained as follows:
\begin{equation}
  P_{coarse}(i,c) = \frac{\exp(s_{i}^c)}{\sum_{k=1}^C \exp(s_{i}^k)}.
  \label{eq10}
\end{equation}

\subsubsection{Dual-Masking Attention Refinement (DMAR)}
The initial patch-level classification scores obtained from Equation~\ref{eq10} often suffer from noise, hindering them from serving as a reliable criterion for class identification~(e.g., leading to false positives for classification in Figure~\ref{fig:ablation_cls}). 
Prior approaches typically utilize pairwise affinity to refine dense classification maps, but require training extra layers~\cite{Ahn2018PSA, Ahn2019IRN}. In contrast, the vision transformer's inherent self-attention mechanism captures the pairwise affinity between patches, allowing us to refine the patch-wise classification scores without incurring additional computational costs. A common way is directly using attention weights from the last few layers~\cite{xu2022mctformer} or all layers~\cite{gao2021tscam} of ViT and performing the refinement as follows:
\begin{equation}
   P_{refined} = \frac{1}{|\psi|}\sum_{l \in \psi} A_l * P_{coarse},
  \label{eq11}
\end{equation}
where $P_{coarse} \in R^{N \times C}$ denotes the coarse score map, $A_l \in R^{N \times N}$ represents the attention weight in the \textit{l-th} layer of ViT, $\psi$ represents the index set of used attention layer and $|\psi|$ is its number of elements. 

However, the affinity captured in the original ViT's MHSA is inaccurate~\cite{ru2022afa}, potentially misleading the refinement process. To address it, we propose a dual-masking strategy, the key idea of which is to neglect unconfident elements in both attention weights $A \in R^{N \times N \times L}$ and coarse score maps $P_{coarse} \in R^{N \times C}$. 
For attention weights, we generate an attention mask~$M_{attn} \in R^{N \times N}$ to select confident elements by leveraging a voting-style approach across all $L$ attention layers. Each confident position should have prominent attention value~(exceeding layer-wise mean value) in at least K layers, which can be represented as: 
\begin{equation}
M_{attn}(i, j) = 1, \, if \, \sum_{l=1}^{L} \mathbb{I}_{(A(i, j, l) > \bar{A_l})}(A) > K,
\label{m_attn}
\end{equation}
where $\mathbb{I}$ is the indicator function, $\bar{A_l}$ is the mean value of $l$-$th$ layer. The refinement procedure is then illustrated as:

\begin{equation}
   \hat{P}_{refined} = \frac{1}{|\psi|}\sum_{l \in \psi} M_{attn} \odot A_l * P_{coarse},
  \label{vote}
\end{equation}
 where $\odot$ denotes the Hadamard product. For coarse score maps, we calculate the average score for each class based on $\hat{P}_{refined}$ and produce a expanded class-wise mask $M_{cls} \in R^{N \times N \times C}$ by ignoring unconfident positions~(below the average score). The final refined scores for each class $c$ can be obtained as follows:
 \begin{equation}
   P_{refined}(c) = \frac{1}{|\psi|}\sum_{l \in \psi} M_{attn} \odot A_l \odot M_{cls}(c) * P_{coarse}(c).
  \label{cls}
\end{equation}

\subsubsection{Class-Wise Reidentification (CWR)}
Although patch-level classification can discover target classes by discriminative local features, it may result in misclassification for the lack of a comprehensive view. 
Therefore, we propose a class-wise reidentification module to further remedy the primary predicted scores for each class from a global view. Specifically, given refined classification scores $P_{refined} \in R^{N \times C}$, we can obtain the confidence of each class $P_{local}$ by corresponding most outstanding patches:
\begin{equation}
   P_{local}(c) = \max_i (P_{coarse}(i, c)),
  \label{eq6}
\end{equation}
For each class, we pick out the highly responsive patches from $P_{refined}$ and form the class-related region (class-wise mask). We crop the image by the bounding box of the region and resize it to a specific size, e.g., $224\times224$. The class-wise mask serves as the attention mask in ViT to exclude patches that do not belong to the class. We input the class-wise image into original CLIP and use [\textit{cls}] token for classification. The obtained global results $P_{global}$ are merged with local scores $P_{local}$ to take advantage of both local and global views. 
\begin{equation}
    P_{final} = \lambda P_{local} + (1 - \lambda) P_{global},
  \label{eq7}
\end{equation}
where $\lambda$ is a coefficient to balance the local and global effect and is simply set to 0.5 in our experiments. Through this fusion process, we can effectively incorporate the valuable insights provided by both local and global views, thereby enhancing the overall classification performance.

\subsection{Appplication on the Downstream Task}
Multi-label classification is a practical task with wide-ranging applications in downstream tasks that rely on image-level labels. In this paper, we explore the use of TagCLIP in conjunction with existing Weakly Supervised Semantic Segmentation (WSSS) methods to tackle annotation-free semantic segmentation. 
Given image-level labels, most WSSS works~\cite{Wang2020SEAM,Xie_2022_CLIMS} leverage Class Activation Mapping (CAM) to find the target class’s related regions in the image and generate segmentation masks based on it. The use of category information provides valuable high-level guidance, enabling WSSS to perform remarkably well, even approaching the performance of fully-supervised settings. We select CLIP-ES~\cite{lin2022clipes} for its outstanding accuracy and efficiency. It is also a training-free framework based on frozen CLIP and more details can be found in~\cite{lin2022clipes}. By leveraging this efficient WSSS method, the whole \textit{classify-then-segment} paradigm requires no dataset-specific training and can realize annotation-free segmentation. We denote this framework as CLS-SEG.

\section{Experiment}
\subsection{Experimental Setup}

\textbf{Dataset and Evaluation Metrics.} To verify the performance of multi-label classification, for fair comparisons, we evaluate our method on PASCAL VOC 2007~\cite{everingham2010pascal} and MS COCO 2014~\cite{lin2014microsoftcoco}. following~\cite{guo2023TAI}. The PASCAL VOC 2007 contains 20 categories and we evaluate on the test set with 4952 images. MS COCO 2014 includes 80 categories, and we take the 40137 images as validation set following the official split.
For downstream semantic segmentation, we conduct experiments on three commonly used datasets, including PASCAL VOC 2012~\cite{everingham2010pascal}, MS COCO 2017~\cite{lin2014microsoftcoco} and COCO-Stuff~\cite{cocostuff}. For Pascal VOC 2012, there are 20 foreground classes, and the remaining pixels are background. The validation set with 1449 images is used for validation. COCO 2017 has 5000 validation images with 80 categories and a background class. COCO-stuff has 4172 validation images of 171 low-level categories. We employ 27 mid-level categories setting following~\cite{shinreco}. Note that our classification framework TagCLIP and segmentation framework CLS-SEG are both training-free and can directly evaluate on the validation set.  We employ mean average precision (mAP) as the evaluation metric for multi-label classification and the mean Intersection over Union (mIoU) for semantic segmentation.

\textbf{Implementation Details.} Our experiments are based on ViT-B/16 pre-trained by CLIP. For multi-label classification, images remain at their original resolution. 
In every operation where a confidence threshold is required, threshold 0.5 is substituted if not otherwise specified, such as thresholds for selecting highly responsive patches in CWR. We adopt the 80 prompts used in CLIP~\cite{CLIP} and background set in~\cite{lin2022clipes}. To determine the potential classes in an image according to classification logits, 
we first perform min-max normalization to scale the logits to $[0, 1]$ and then set 0.5 to determine positive categories.

\begin{table}
  \begin{center}
  \resizebox{0.92\columnwidth}{!}{
  \begin{tabular}{lccc}
    \toprule
    Method  & Extra Training Data   &  VOC & COCO \\
    \midrule
    \multicolumn{4}{l}{\textbf{\textit{Supervised specialist:}}} \\
    SARB  & 10\% Data        & 83.5  & 75.5\\
    DualCoOp  & 10\% Data        & 90.3  & 78.7\\
    TAI-DPT  & 10\% Data        & 93.3  & 81.5\\
    \hline
    \multicolumn{4}{l}{\textbf{\textit{Open-vocabulary generalist:}}} \\
    TAI-DPT  & COCO captions        & 88.3  & 65.1\\
    CLIP$^\dagger$  & None        & 79.5  & 54.2\\
    CLIP  & None        & 85.8  & 63.3\\
    DPT$^\dagger$  & None        & 83.4  & 59.6\\
    DPT  & None        & 86.2  & 64.3\\
    CLIPSurgery  & None        & 85.4  & 61.2\\
    \textbf{TagCLIP(Ours)}  & None        & \textbf{92.8}  & \textbf{68.8}\\
    \bottomrule
  \end{tabular}
  }
  \end{center}
  \caption{Experimental results of multi-label classification. $^\dagger$ represents not using softmax on classification scores.}
  \vspace{-2mm}
  \label{tab:cls}
\end{table}

\subsection{Experimental Results}
\subsubsection{Multi-label classification.}
To demonstrate the effectiveness of our proposed TagCLIP, we compare it with other CLIP-based approaches. Some supervised specialist methods leverage partial data on downstream datasets to train customized models,  
including SARB~\cite{pu2022sarb}, DualCoOp~\cite{sun2022dualcoop}, TAI-DPT~\cite{guo2023TAI}. The use of downstream data limits their generalization. Another training-based manner has no access to downstream data but trains on curated caption data, which enables arbitrary category recognition.
The others are merely based on frozen CLIP and thus inherit its outstanding generalization capability, including CLIP~\cite{CLIP}, DPT~\cite{guo2023TAI}, CLIPSurgery~\cite{li2023clipsurgery}. 

\begin{table}
  \begin{center}
  \resizebox{0.88\columnwidth}{!}{
  \begin{tabular}{lccc}
    \toprule
    Method  & VOC & COCO & COCO-Stuff \\
    \midrule
    \multicolumn{4}{l}{\textbf{\textit{Vanilla USS methods}}} \\
    IIC       & 9.8    & - & 6.7  \\
    MaskContrast  & 35.0  & 3.73 & -   \\
    TransFGU & 37.2 & 12.7 & 17.5\\
    MaskDistill  & 45.8  & - & -     \\
    PiCIE & - & - & 13.8 \\
    PiCIE+H & - & - & 14.4 \\
    \midrule
    \multicolumn{2}{l}{\textbf{\textit{CLIP-based methods}}} \\
    MaskCLIP$^\ddagger$  & 42.1  & 20.2 & 23.9      \\
    CLIPSurgery$^\ddagger$  & 41.5  & 25.2 & 29.7      \\
    GroupViT  & 52.3 & 24.3 & -       \\
    SegCLIP  & 52.6 & 26.5 & -       \\
    ReCo & 34.2 & 17.1 & 26.3 \\
    NamedMask  & 59.2  & 27.7 & -      \\
    \textbf{CLS-SEG (Ours)}    & \textbf{64.8} & \textbf{34.0} & \textbf{30.1} \\
    \textbf{CLS-SEG$^*$(Ours)}    & \textbf{68.7} & \textbf{35.3} & \textbf{31.0} \\
    \bottomrule
  \end{tabular}
  }
  \end{center}
  \caption{Results of annotation-free semantic segmentation. The vanilla USS results are based on K-means clustering. $^\ddagger$ represents we re-implement it with the same experimental setting as ours. $^*$ means using denseCRF to postprocess.}
  \vspace{-2mm}
  \label{tab:seg}
\end{table}

In Table~\ref{tab:cls}, $^\dagger$ represents directly treating logits before softmax as classification scores~\cite{guo2023TAI} because these logits can reflect the similarity between image and text features. We find that there is a major performance degradation without softmax activation, which may stem from the use of contrastive loss during pre-training of CLIP. Results in Table~\ref{tab:cls} demonstrate that our proposed framework performs surprisingly well. It enhances the multi-label classification performance of original CLIP by a large margin, i.e., 7.0\% and 5.5\% on VOC and COCO, respectively. Our method surpasses all works that require no extra training data on both VOC and COCO. It also compares favorably with the works requiring extra data and training. More experiment results are available in Appendix.

\subsubsection{Segmentation performance.}
We provide our annotation-free segmentation result with tags generated by TagCLIP as pseudo labels and compare them with both \textbf{vanilla USS methods} (including IIC~\cite{ji2019iic}, MaskContrast~\cite{van2021maskcontrast}, TransFGU~\cite{yin2022transfgu}, MaskDistill~\cite{van2022maskdistill}, PiCIE(+H)~\cite{cho2021picie}) and recent \textbf{CLIP-based works} (including  MaskCLIP~\cite{zhou2022maskclip}, CLIPSurgery~\cite{li2023clipsurgery}, GroupViT~\cite{xu2022groupvit}, SegCLIP~\cite{luo2023segclip}, ReCo~\cite{shinreco}, NamedMask~\cite{shin2022namedmask}) in Table~\ref{tab:seg}. 

\begin{figure}[t]
  \centering
   \includegraphics[width=0.95\linewidth]{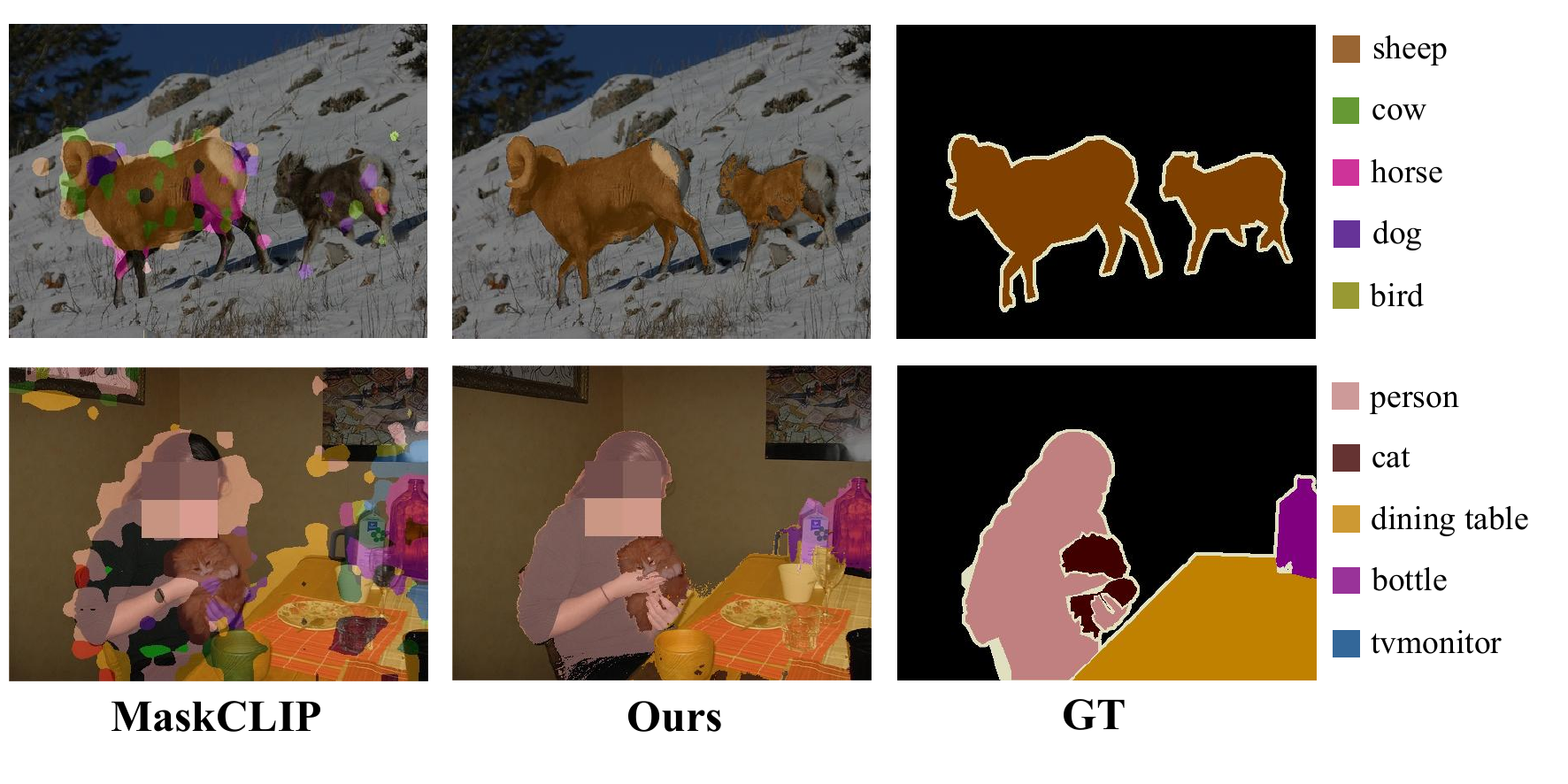}
   \caption{Visualizations of segmentation results for MaskCLIP~\cite{zhou2022maskclip} and ours. MaskCLIP has more false positives for the lack of category information.
   }
   \label{fig:maskclip_ours}
   \vspace{-2mm}
\end{figure}

We observe that our CLS-SEG outperforms vanilla USS and other CLIP-based methods by a large margin on all three datasets, which demonstrates the high quality of our generated tags and verifies the effectiveness of this \textit{classify-then-segment} paradigm. From Figure~\ref{fig:maskclip_ours}, we find that high-level conceptual guidance provided by category information in an image is essential to obtain high-quality segmentation masks because: 1) it prevents false predictions caused by confusing textures among semantically similar classes, e.g., the skin of the cow and sheep;  
2) it can comprehensively identify some categories with large intra-class variance, e.g., different parts of a person can be identified as a whole with superior semantic concepts. The results indicate that classification helps segmentation and may provide inspiration for future research.

\begin{figure}[t]
  \centering
   \includegraphics[width=0.95\linewidth]{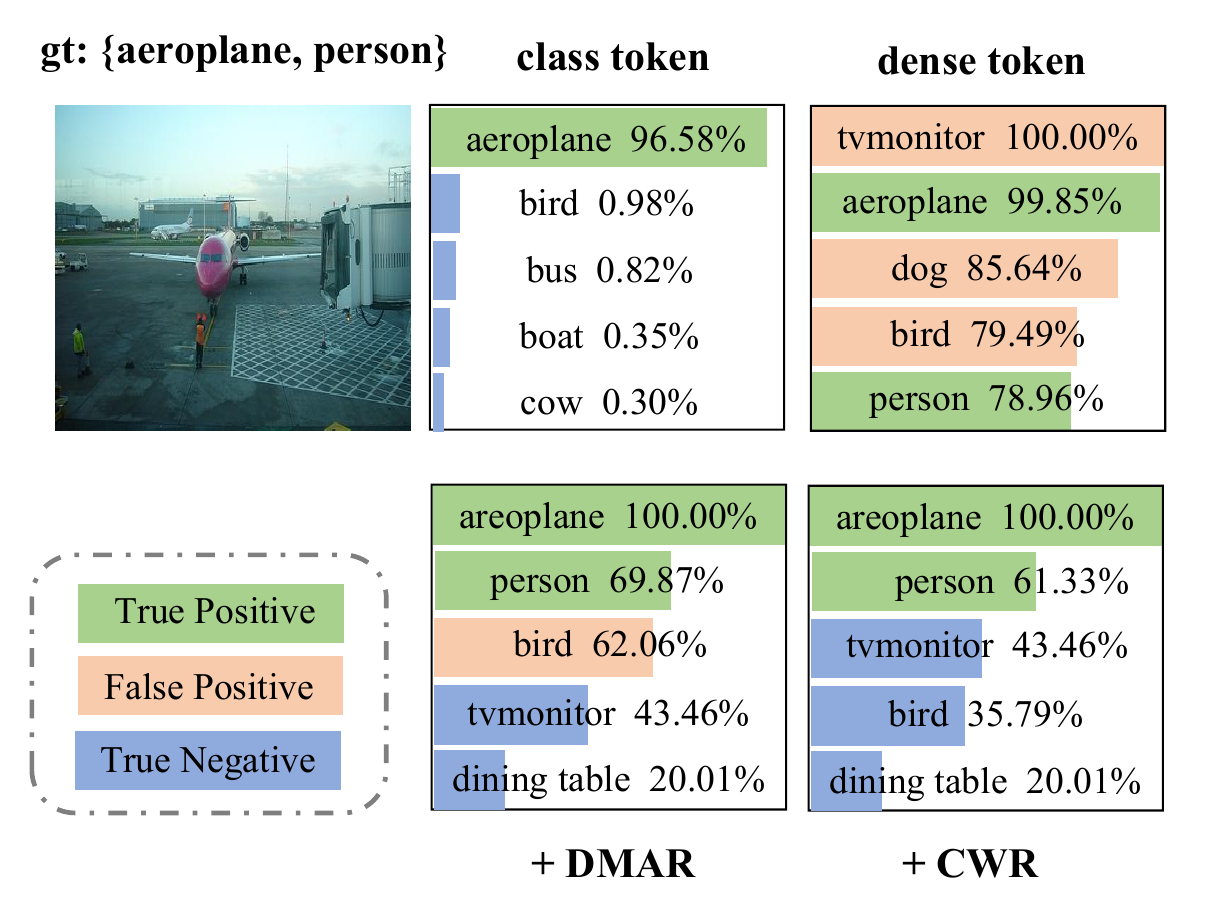}
    \vspace{-2mm}
   \caption{The classification results of \textit{class} token and \textit{dense} token and the effect of our proposed modules. We use 0.5 as the threshold by default.
   }
   \label{fig:ablation_cls}
   \vspace{-2mm}
\end{figure}

\begin{table}[t]
  \begin{center}
  \resizebox{0.9\columnwidth}{!}{
  \begin{tabular}{ccc|cc}
    \toprule
    Coarse Score  & DMAR   &  CWR & mAP  & mIoU  \\
    \midrule
    \Checkmark       &         &   &  85.4  & 30.9 \\
    \Checkmark       &         & \Checkmark  &  88.0  & 55.2 \\
    \Checkmark  & \Checkmark        &   & 93.9  & 63.7 \\
    \Checkmark  & \Checkmark        & \Checkmark  & 94.1 & 64.8\\
    \bottomrule
  \end{tabular}
  }
  \end{center}
  \caption{Results for the effectiveness of DMAR and CWR module in terms of classification and semantic segmentation. The results are evaluated on the PASCAL VOC 2012 validation set.}
  \label{tab:ablation1}
\end{table}

\begin{figure}[t]
    \centering
    \subfigbottomskip=2pt
    \subfigcapskip=-5pt
    \subfigure{
    \includegraphics[width=0.45\linewidth]{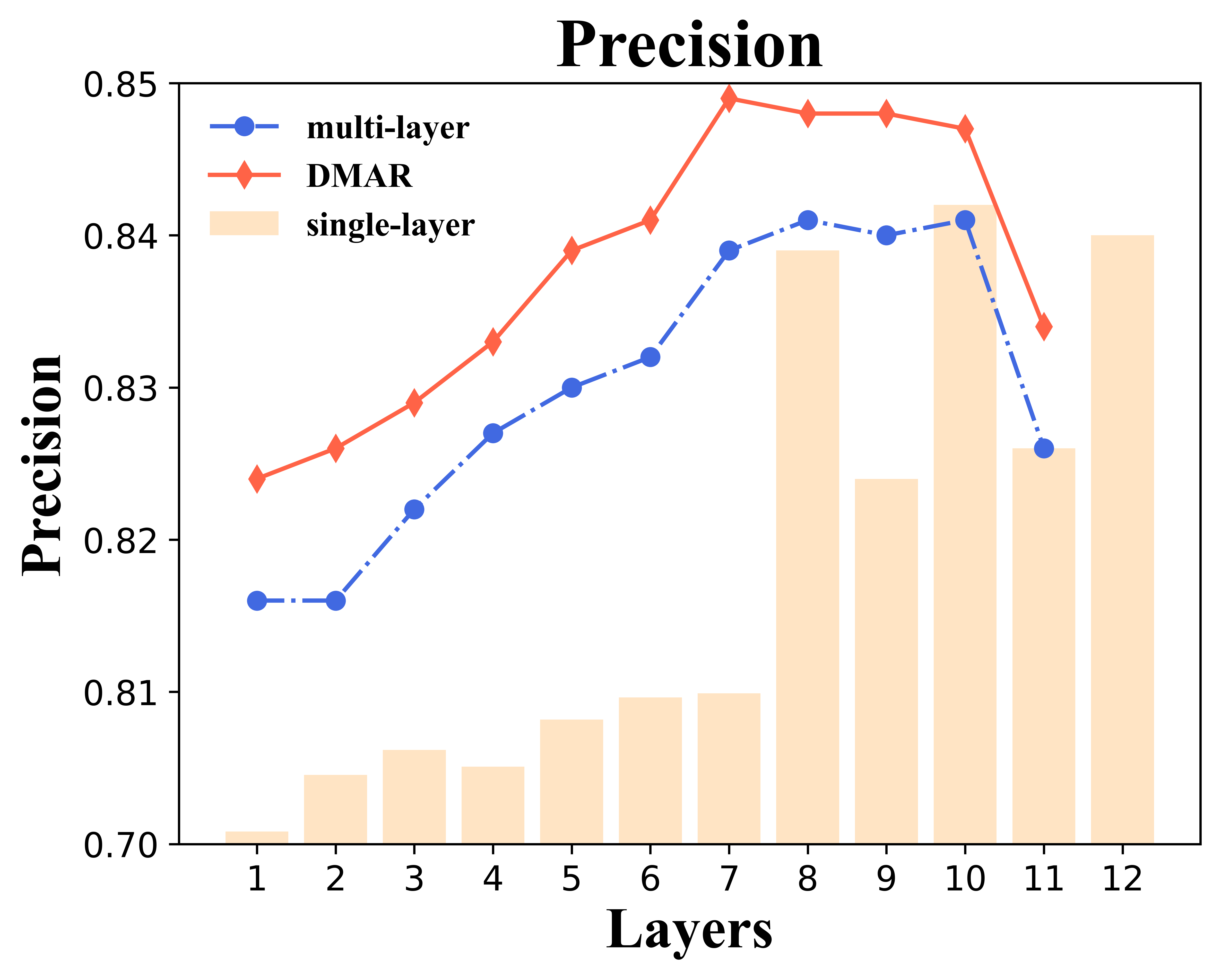}
    \label{fig1:a}
    }
    \subfigure{
    \includegraphics[width=0.45\linewidth]{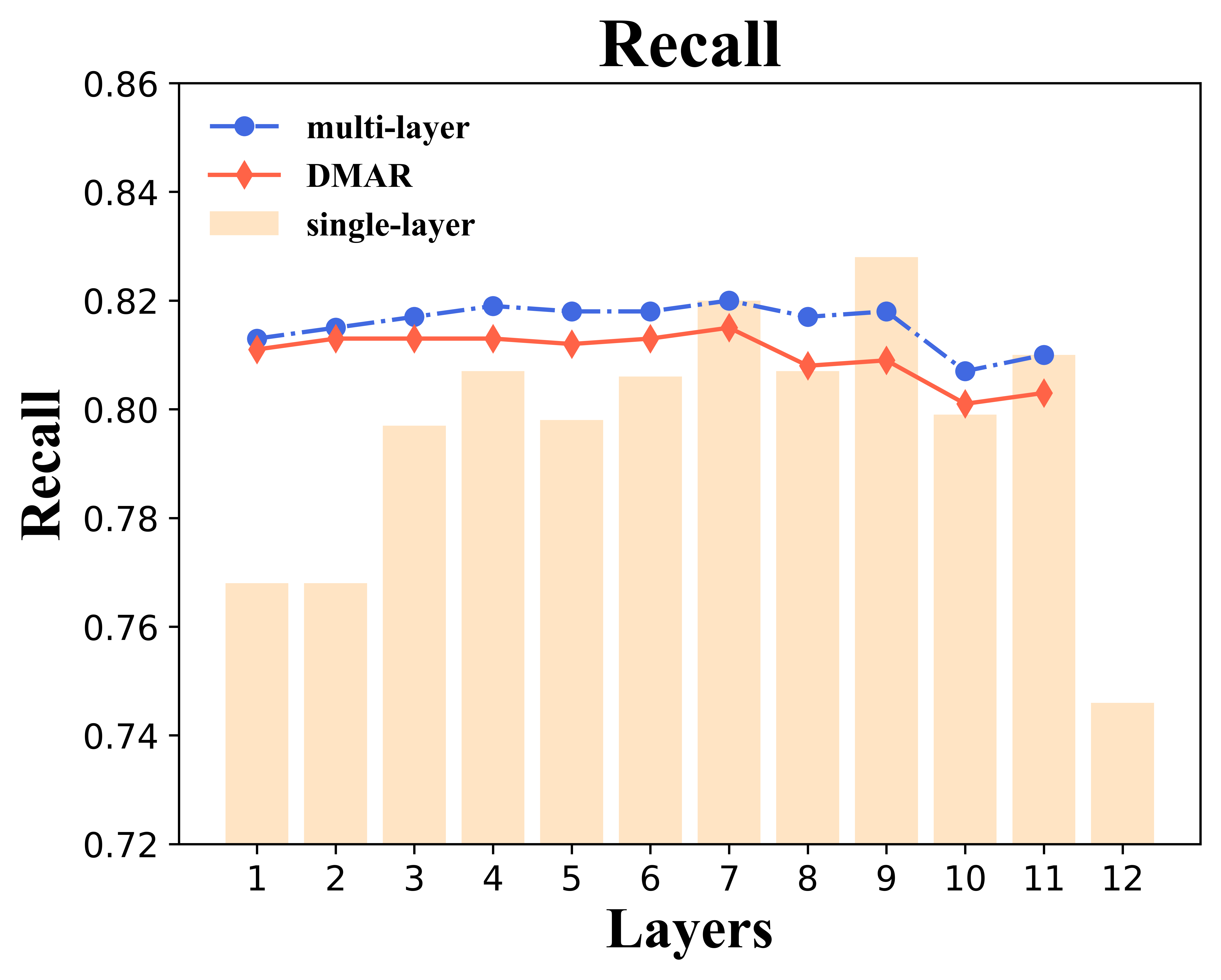}
    \label{fig1:b}
    }
    \\
    \subfigure{
    \includegraphics[width=0.45\linewidth]{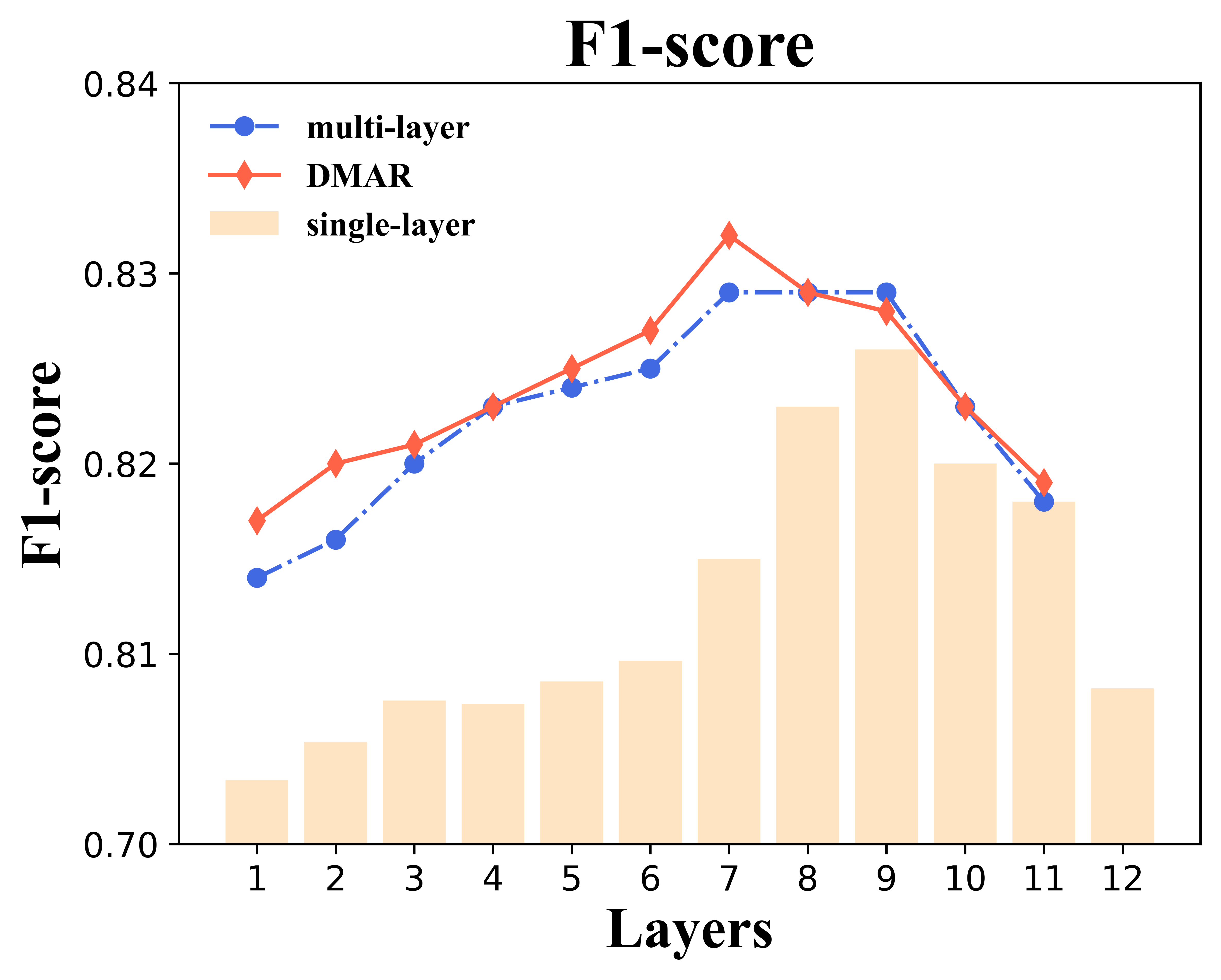}
    \label{fig1:c}
    }
    \subfigure{
    \includegraphics[width=0.45\linewidth]{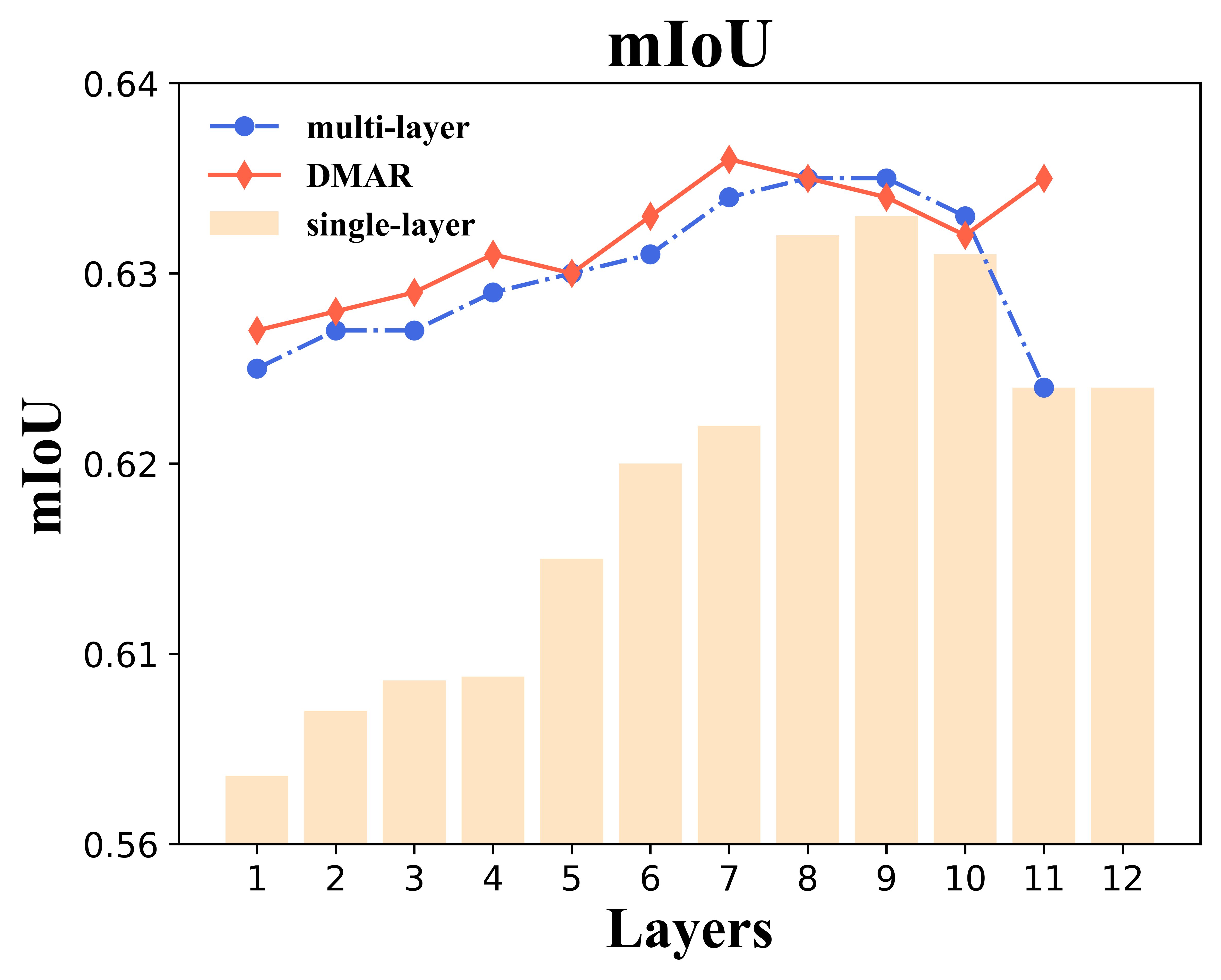}
    \label{fig1:d}
    }
    \caption{Comparison of single-layer and multi-layer attention refinement in terms of classification and segmentation tasks. For the single-layer setting, each tick \textit{i} on the x-axis represents merely adopting attention weight in \textit{i-th} layer. For the multi-layer setting, the \textit{i-th} x-tick means fusing \textit{i-th to 11-th} layers attention weights to refine coarse classification scores. We rule out the last attention layer during fusing.}
    \label{fig:attn_ablation}
\end{figure}

\subsection{Ablation Study}
\textbf{Effect of DMAR and CWR.}
In Table~\ref{tab:ablation1}, we evaluate the effect of DMAR and CWR in terms of classification and segmentation. The DMAR can refine the coarse scores remarkably and CWR can further boost the performance. We provide a qualitative case in Figure~\ref{fig:ablation_cls}. After DMAR, most irrelevant categories can be suppressed. The CWR can coordinate with DMAR to double-check the refined scores from a global view. Therefore, the scores of false positives and false negatives can be suppressed and improved, respectively.

\textbf{Effect of attention layers used in DMAR module.}
To determine the appropriate attention layers in CLIP-ViT for classification score refinement, we first compare single-layer attention weight with multi-layer in terms of classification (\textit{precision, recall and f1-score}) and segmentation (\textit{mIoU}) performance. From Figure~\ref{fig:attn_ablation}, we can draw the following conclusions: 1) Fusing multi-layer attention weights usually performs better and more robustly than single-layer. 2) The performance of the first few attention layers is unsatisfactory, which mainly stems from the weak attention and features these layers learned. 3) The last attention layer is inaccurate among the last few layers, which corresponds to our analysis above. We also present the performance of our proposed dual-masking strategy, which effectively mitigates the impact of noise and improves original attention refinement in most cases. This strategy demonstrates significant precision gains with only a slight recall drop, leading to better classification and segmentation performance in general. Based on these observations, we fuse the last four attention weights except the last one in our experiments.

\section{Conclusion}

This paper proposes TagCLIP, a simple and effective framework designed to enhance the multi-label classification capability of the original CLIP model. It follows a local-to-global paradigm and consists of three key steps: patch-level classification, dual-masking attention refinement (DMAR), and class-wise reidentification (CWR). Benefiting from these steps, TagCLIP unlocks the potential of CLIP and can serve as a generalizable annotator that provides high-quality image tags without dataset-specific training. Additionally, we validate the practicality of treating generated tags as pseudo labels for the downstream weakly supervised semantic segmentation (WSSS) task and find this \textit{classify-then-segment} paradigm surpasses previous bottom-up style annotation-free segmentation methods by a large margin. This demonstrates the effectiveness and versatility of TagCLIP and highlights its potential in various downstream applications.

\section{Acknowledgements}
This work was supported in part by The National Nature Science Foundation of China (Grant Nos: 62273303, 62273301, 62273302, 62036009, 61936006), in part by Ningbo Key R\&D Program (No.2023Z231, 2023Z229), in part by Yongjiang Talent Introduction Programme (Grant No: 2022A-240-G), in part by the Key R\&D Program of Zhejiang Province, China (2023C01135), in part by the National Key R\&D Program of China (NO.2022ZD0160100).

\bibliography{aaai24}

\newpage

\appendix
\renewcommand\thefigure{S\arabic{figure}}
\renewcommand\thetable{S\arabic{table}}

\setcounter{figure}{0}
\setcounter{table}{0}

\vspace{-4mm}

\begin{strip}
\begin{center}
  \begin{tabular}{l}
  \LARGE \textbf{Appendix} \\
  \end{tabular}
\end{center}
\end{strip}


\begin{figure}[t]
  \centering
   \includegraphics[width=0.6\linewidth]{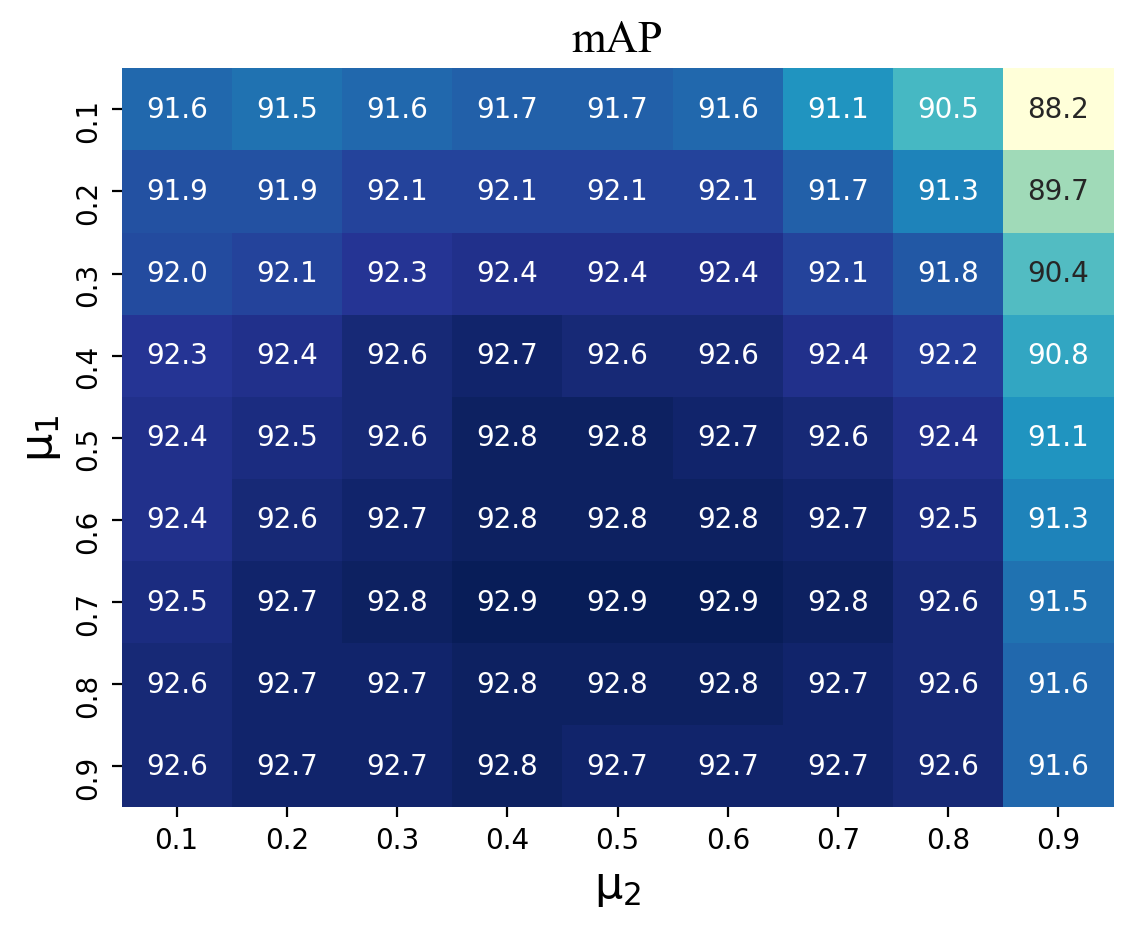}

   \caption{Analysis of $\mu_1, \mu_2$ on VOC 2007 test set.}
   \label{fig:mu}
   \vspace{-2mm}
\end{figure}

\section{Additional Experiments}
\subsection{Effect of Parameters in CWR}
In CWR module, some thresholds are introduced to select potential categories ($\mu_1$) and highly responsive patches ($\mu_2$). 
We analyze the effect of $\mu_1, \mu_2$ in Figure~\ref{fig:mu}. Results show that the classification performance is not sensitive to these parameters with moderate values. We didn't specifically select them and simply set 0.5 in our paper.

\subsection{Effect of Fusion Coefficient $\lambda$}
\begin{figure}[t]
  \centering
   \includegraphics[width=0.62\linewidth]{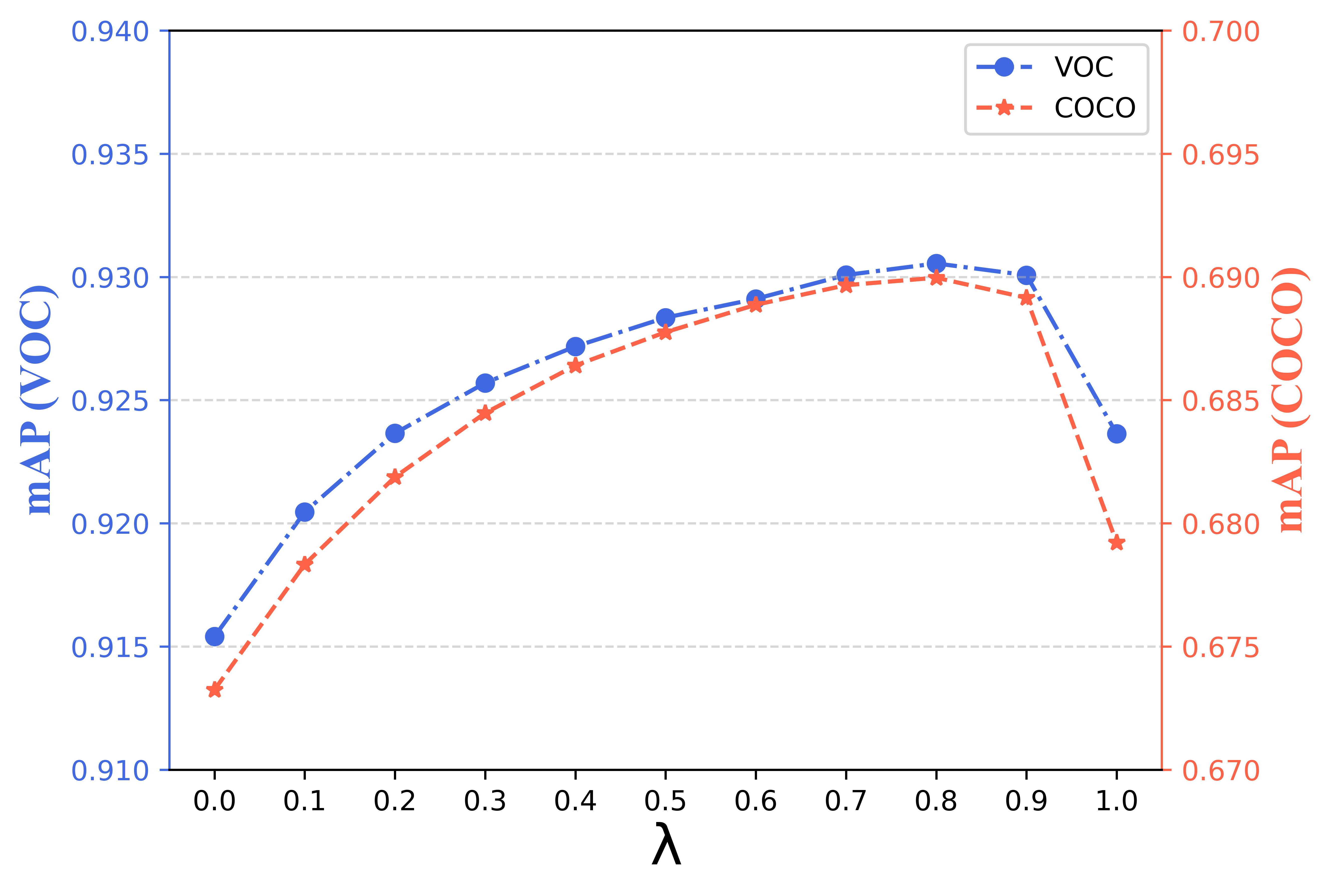}

   \caption{Effect of $\lambda$ on the classification performance on PASCAL VOC 2007 test set and COCO 2014 val set.}
   \label{fig:lambda}
   \vspace{-2mm}
\end{figure}

In our local-to-global framework, we fuse the local and global classification scores leveraging a coefficient $\lambda$, which controls the balance between the local and global effect. We analyze the variation of classification performance using different $\lambda$ in Table~\ref{fig:lambda}. The results demonstrate that fusing local and global features can boost classification performance that solely using local features~($\lambda=1$) or global features~($\lambda=0$), which validates the effectiveness of our framework. We did not specifically select the best $\lambda$ but simply set it to 0.5 in our experiments.

\subsection{Effect of Background Set}
We conduct experiments on both object-centric PASCAL VOC~\cite{everingham2010pascal}, MS COCO~\cite{lin2014microsoftcoco} and scene-centric COCO-Stuff~\cite{cocostuff} benchmarks. However, the object-centric datasets only include annotations for specific objects and the remainings are treated as background, which is not beneficial to our patch-level classification. We follow CLIP-ES~\cite{lin2022clipes} to leverage some commonly used background categories. Specifically, the background set used in PASCAL VOC is \{\textit{ground, land, grass, tree, building, wall, sky, lake, water, river, sea, railway, railroad, helmet, cloud, house, mountain, ocean, road, rock, street, valley, bridge, sign, keyboard}\}. For COCO, the last two background categories are removed because similar categories have been defined in the dataset. No background set is used for COCO-Stuff because the defined 171 categories have covered both things and stuff categories. 
We present the effect of the background set on the classification and segmentation task in Table~\ref{tab:bg}. It contributes to the overall performance.

\begin{table}
  \begin{center}
  \begin{tabular}{ccc}
    \toprule
    Background Set  & VOC   &  COCO  \\
    \midrule
    \XSolidBrush    &    92.4  & 67.9 \\
    \Checkmark    &      92.8  & 68.8 \\
    
    \bottomrule
  \end{tabular}
  \end{center}
  \caption{Results for the effect of background set on multi-label classification task~(mAP). The results are based on PASCAL VOC 2007 test set and COCO 2014 val set.}
  \label{tab:bg}
  \vspace{-2mm}
\end{table}


\section{Limitations}
The proposed TagCLIP follows a local-to-global framework to enhance the multi-label classification of CLIP. It demonstrates strong performance on common multi-label datasets (like VOC and COCO) where the defined categories are mutually competitive, e.g., \textit{cat and dog}. However, TagCLIP may not be optimal for datasets with inclusion relationships between categories, e.g., \textit{cat and animal}. This is because the hierarchy-style categories are not well-suited for pre-trained CLIP, which is based on contrastive loss and fosters competition among different classes. A potential solution is adopting the self-training approach.
Besides, some scene categories (like \textit{sky}) may not benefit significantly from discriminative local cues and require stronger conditions for accurate recognition. 
Consequently, our method performs significantly well for object-centric datasets but there is room for improvement when facing hierarchic categories with different granularity. We take solving this issue as future research.


\section{Additional Qualitative Results}
\begin{figure*}
  \centering
   \includegraphics[width=0.95\linewidth]{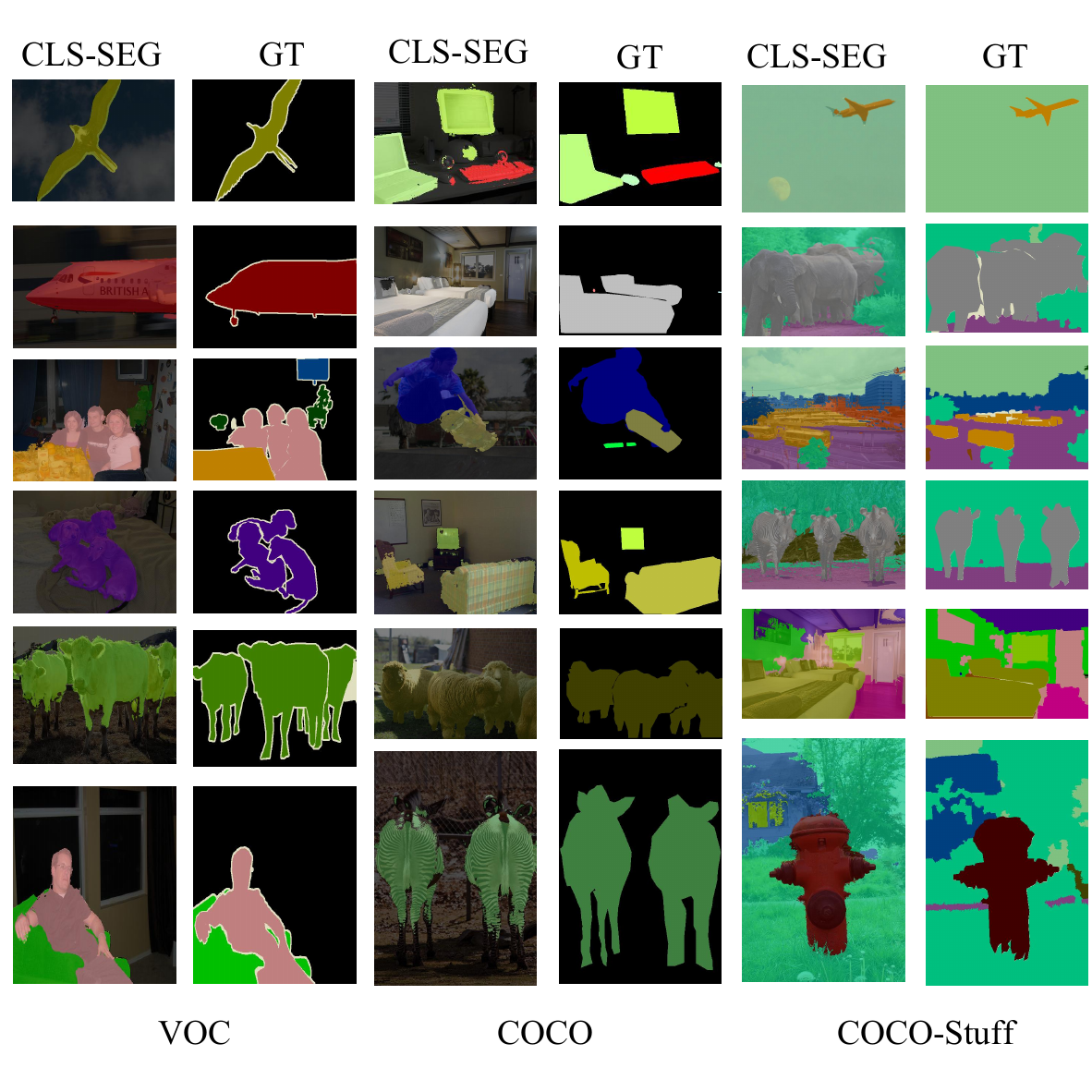}

   \caption{Additional Visualizations of our method on VOC, COCO and COCO-Stuff.}
   \label{fig:more_vis}
\end{figure*}

In Figure~\ref{fig:more_vis}, we provide more visualizations leveraging our CLS-SEG framework on different benchmarks. Our method performs well on these challenging datasets, especially for rigid objects and animals. There is still room for improvement in rare classes and stuff categories, which may be solved by more exact category descriptions.

\end{document}